\documentclass[aps,pra,reprint,longbibliography]{revtex4-1}

\usepackage{graphicx}
\usepackage{amsmath}
\usepackage{amssymb}
\usepackage{textcomp}

\begin{document}
	
\title[]{Superconducting Optoelectronic Neurons \uppercase\expandafter{\romannumeral 4 \relax}: Transmitter Circuits}
	
\author{Jeffrey M. Shainline, Adam N. McCaughan, Sonia M. Buckley, Richard P. Mirin, and Sae Woo Nam}
\affiliation{National Institute of Standards and Technology, 325 Broadway, Boulder, CO, 80305}		
	
\author{Amir Jafari-Salim}
\affiliation{HYPRES, Inc., 175 Clearbrook Rd, Elmsford, NY 10523, USA}		
	
\date{\today}
	
\begin{abstract}
A superconducting optoelectronic neuron will produce a small current pulse upon reaching threshold. We present an amplifier chain that converts this small current pulse to a voltage pulse sufficient to produce light from a semiconductor diode. This light is the signal used to communicate between neurons in the network. The amplifier chain comprises a thresholding Josephson junction, a relaxation oscillator Josephson junction, a superconducting thin-film current-gated current amplifier, and a superconducting thin-film current-gated voltage amplifier. We analyze the performance of the elements in the amplifier chain in the time domain to calculate the energy consumption per photon created for several values of light-emitting diode capacitance and efficiency. The speed of the amplification sequence allows neuronal firing up to at least 20\,MHz with power density low enough to be cooled easily with standard $^4$He cryogenic systems operating at 4.2\,K.
\end{abstract}
	
%\keywords{neural networks, neuromorphic computing, integrated photonics, superconducting electronics}
	
\maketitle

%\tableofcontents
	
\section{\label{sec:introduction}Introduction}
A physical substrate for cognition must enable integration of information across multiple spatial and temporal scales \cite{to2004,bu2006}. We have argued that light is ideal for communication in neural systems \cite{shbu2017,sh2018a}. In the previous papers in this series \cite{sh2018b,sh2018c} we have designed superconducting circuits that receive photonic signals and convert them into an integrated supercurrent. Such synaptic receivers must partner with a light source to generate a neuronal firing event when the integrated current reaches a threshold. In Ref.\,\onlinecite{sh2018b} we propose a Josephson junction (JJ) could be used as a thresholding element. In this work, we show that the flux quantum generated by a thresholding JJ can trigger an amplification sequence resulting in an optical pulse containing one to 100,000 photons. These superconductor/semiconductor hybrid circuits achieve electrical-to-optical transduction and facilitate communication with high fanout and light-speed delays in complex networks of superconducting optoelectronic neurons. We refer to these amplifier circuits as the transmitter of the neuron. Connectivity between these neurons via dielectric waveguides is discussed in Ref.\,\onlinecite{sh2018e}.

\begin{figure} %[t] %[htb]
	\centerline{\includegraphics[width=8.6cm]{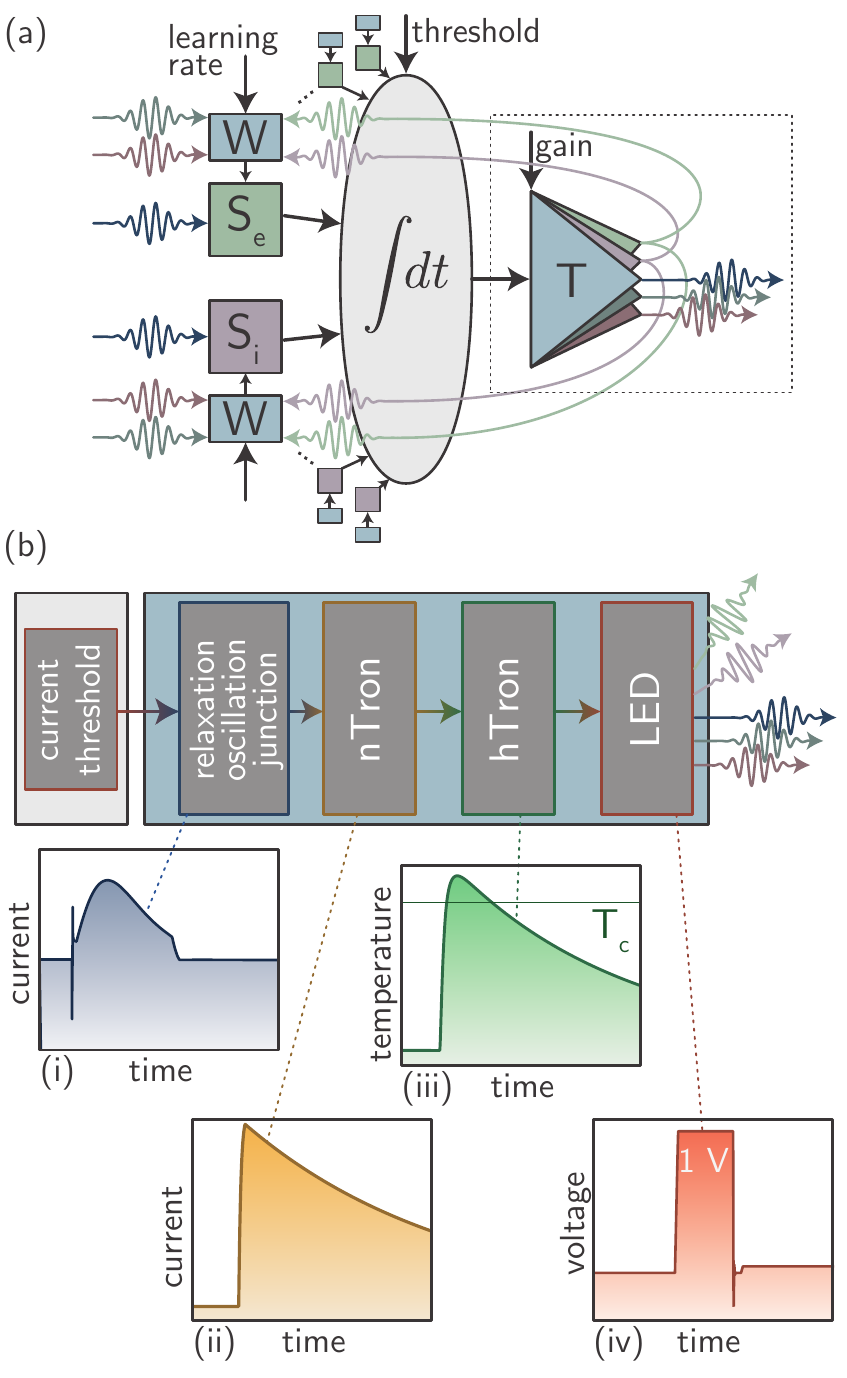}}
	\caption{\label{fig:transmitters_schematic}(a) Schematic of the neuron showing excitatory ($\mathsf{S_e}$) and inhibitory synapses ($\mathsf{S_i}$) connected to an integrating component with a variable threshold. The wavy, colored arrows are photons, and the straight, black arrows are electrical signals. The output of the thresholding integrator is input to the transmitter, $\mathsf{T}$. The dashed box encloses the transmitter that is the focus of this work. (b) Sequence of events during neuronal firing event. (i) Current threshold is reached in the neuronal thresholding loop, causing the relaxation oscillation junction to produce transient current to the nTron gate. (ii) Current from the relaxation oscillator junction drives the gate of the nTron normal, causing the nTron channel current to be diverted to the gate of the hTron. (iii) The current from the nTron through the gate of the hTron drives the channel of the hTron normal, resulting in a voltage pulse across the hTron. (iv) The LED is in parallel with the hTron, so the voltage across the hTron results in a voltage across the LED. This voltage is sufficient to forward-bias the $p-n$ junction to produce light.}
\end{figure}
The superconducting optoelectronic neurons described in this series of papers are referred to as loop neurons. The loop neuron in which this transmitter operates is shown in Fig.\,\ref{fig:transmitters_schematic}(a). Operation is as follows. Photons from afferent neurons are received by single-photon detectors (SPDs) at a neuron's synapses. Using Josephson circuits, these detection events are converted into an integrated supercurrent that is stored in a superconducting loop. The amount of current added to the integration loop during a synaptic photon detection event is determined by the synaptic weight. The synaptic weight is dynamically adjusted by another circuit combining SPDs and JJs. When the integrated current from all the synapses of a given neuron reaches a threshold, an amplification cascade begins, resulting in the production of light from a waveguide-integrated light-emitting diode (LED). The photons thus produced fan out through a network of passive dielectric waveguides and arrive at the synaptic terminals of other neurons where the process repeats. The transmitter circuit which receives the threshold signal and produces a pulse of light is the subject of this paper.
	
\section{\label{sec:conceptualOverview}Conceptual Overview}
A large-scale neural system must achieve excellent communication and exceptional power efficiency. Optical signals are ideal for communication, and superconducting detectors enable power efficiency. A central technical challenge in designing superconducting optoelectronic hardware is to produce optical signals at telecommunication wavelengths with superconducting electronic circuits. The superconducting energy gap \cite{ti1996} is in the millivolt range, making it difficult for superconducting circuits to produce the one volt needed to appreciably alter the carrier concentrations in semiconductor electronic and optoelectronic devices. This voltage mismatch makes it difficult for superconducting electronic circuits to interface with CMOS logic \cite{ka1999} and memory \cite{vafe2002}. 

A common approach to increase voltage in superconducting circuits is to place JJs in series \cite{suin1988}. The achievable voltage scales as the superconducting energy gap multiplied by the number of junctions. Order one thousand junctions must be utilized to drive the light sources we intend to employ \cite{shbu2017,buch2017,bu2018}. While superconducting circuits operate at low voltage, they can sustain plenty of current. Supercurrents can be converted to voltage with a resistor. A microamp across a megaohm produces a volt. A small meandering wire of a superconducting thin film can easily produce a megaohm resistance in the normal-metal state. Thus, we can convert a current-biased superconducting wire into a voltage source by switching the wire between the superconducting and normal states.

Such a voltage source is not commonly used in superconducting electronics because it tends to be slow and consume more energy per operation than a JJ. The advantages of superconducting electronics for digital computing are largely speed and efficiency, while the advantages of superconducting optoelectronic networks are largely \textit{communication} and efficiency. While superconducting electronics aspire to operation above 100\,GHz, superconducting optoelectronic networks comprising neurons firing up to 20 MHz would be a radical increase relative to the maximum frequency of 600\,Hz in biological neural systems \cite{stsa2000,budr2004,bu2006}. The goal of neural rather than digital computing liberates us to use superconducting devices slower than JJs for events that only occur once per neuronal firing. The speed and efficiency of JJs are still leveraged by loop neurons to distinguish synaptic efficacy during each synaptic firing event \cite{sh2018b}. Thus, we can use a switching element leveraging the superconductor-to-normal phase transition in a neural context, provided it operates every neuronal firing event and not every synaptic firing event. 

An effective means of breaking the superconducting phase is to heat the wire locally. Thermal devices are generally slow and power hungry, but the small volume, small temperature change, and small heat capacity at low temperature \cite{lasa1988,du2015} enable switching times on the order of a nanosecond with femtojoules of energy. While such an amplifier is not suitable for the high speed and low power of flux-quantum logic \cite{hehe2011,mu2011,li2012,taoz2013}, a single firing of the switch can produce thousands of photons, making it very useful and efficient in this neural context.

This high-impedance, phase-change switch is called an hTron \cite{zhto2018}. The heating operation required to switch the hTron from the superconducting to metal state can be achieved through dissipation in a Joule heater. Fairly large current is required to provide sufficient power. While the hTron is equipped to deliver voltage to drive the semiconductor light source, the thresholding JJ \cite{sh2018b} will provide only fluxons when the neuron's threshold is reached. These fluxons are insufficient to thermally switch the state of the hTron. An intermediate current amplifier is therefore required to convert fluxons from the thresholding JJ into current across the Joule heating element of the hTron. For this step of the amplification process, an nTron \cite{mcbe2014} will suffice. The nTron is a three-terminal thin-film superconducting current amplifier. A small current in the constricted gate can exceed the local critical current and drive the path from source to drain normal. An impedance on the order of a kilohm can be produced quickly, and the current from the source to drain will be diverted to a load. In the present case, that load is the $\approx$\,10\,$\Omega$ resistive element of the hTron. 

A summary of the circuit operation is shown in Fig. \ref{fig:transmitters_schematic}. The loop neuron of Fig.\,\ref{fig:transmitters_schematic}(a) is the subject of this series of papers, and the transmitter circuit enclosed in the dashed box is the subject of this paper. The sequence of events in the operation of the transmitter is shown in Fig.\,\ref{fig:transmitters_schematic}(b). The neuronal thresholding loop described in Ref.\,\onlinecite{sh2018b} reaches threshold due to integrated current from synaptic firing events, and the thresholding JJ produces one or more fluxons. These fluxons enter the gate of the nTron, exceeding the gate critical current. The nTron source/drain current is then diverted to the hTron gate. Joule heating in the hTron gate produces a temperature shift of several kelvin, switching the hTron to the high-impedance state. The hTron source-drain current suddenly experiences 1\,M$\Omega$, and the current rapidly switches to first charge the LED capacitor and then, once sufficient voltage is achieved across the capacitor, current is driven through the diode. This current generates photons through electron-hole recombination. These photons are the optical source for neural communication.

\begin{figure} %[t] %[htb]
	\centerline{\includegraphics[width=8.6cm]{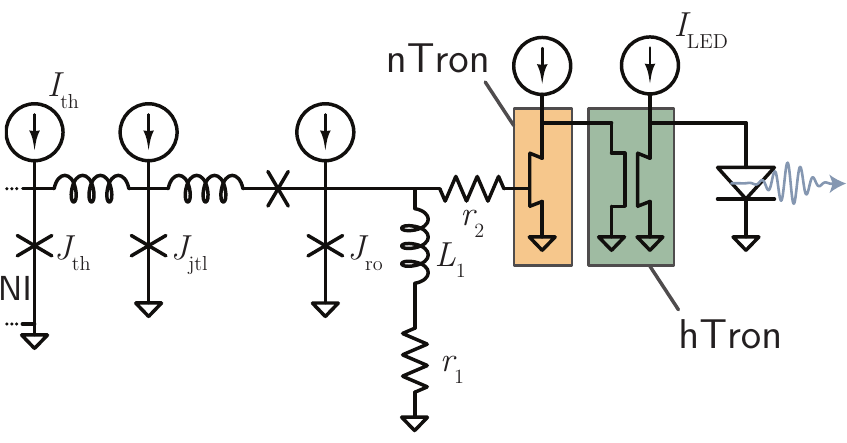}}
	\caption{\label{fig:transmitters_amplifierChain_circuit}Circuit diagram of the amplifier chain under consideration showing the thresholding JJ, voltage-state JJ, nTron, hTron, and LED.}
\end{figure}
A circuit diagram of the amplifier chain is shown in Fig.\,\ref{fig:transmitters_amplifierChain_circuit}. This circuit is relatively complex and uses a significant amount of power. It may be possible to develop a more elegant solution to the problem. Any solution must meet several operating criteria: 1) The transmitter must threshold on a low-current signal from the neuronal thresholding (NT) loop; 2) The amplifier chain must convert this signal to the voltage necessary to produce light from a semiconductor diode; 3) The operation should happen at least as fast as the recovery of the single-photon detectors in the loop neuron synapses (a few tens of nanoseconds); 4) A number of photons appropriate to communicate with the neuron's synaptic connections must be produced; 5) The number of photons created must be dynamically variable with a bias current; 6) The energy of a firing event must be low enough that an ensemble of neurons in realistic network operation have power density low enough that heat can be removed with standard cryogenic systems or when immersed in liquid helium; 7) Total power consumption must be low enough to enable scaling to neural systems with billions of interacting nodes with firing rates up to 20\,MHz. The transmitter circuits presented here satisfy all these criteria.

We begin with the design of the hTron driving the LED and work backward through the circuit of Fig.\,\ref{fig:transmitters_amplifierChain_circuit}.
	
\section{\label{sec:LED}Driving the light-emitting diode}
The operation of the amplifier chain shown in Fig. \ref{fig:transmitters_amplifierChain_circuit} has been described qualitatively in Sec.\,\ref{sec:conceptualOverview}, but to reach a quantitative design we must ensure the voltage required to drive the LED can be sustained for a duration commensurate with the number of photons required. This duration will depend on the drive current, the capacitance of the LED, the efficiency of the LED, and the number of synaptic connections served by the neuron. Once the drive requirements of the LED are understood, we can proceed to design an hTron that can meet these drive requirements. The first task is to explore the operation of the LED.

\begin{figure} %[t] %[htb]
	\centerline{\includegraphics[width=8.6cm]{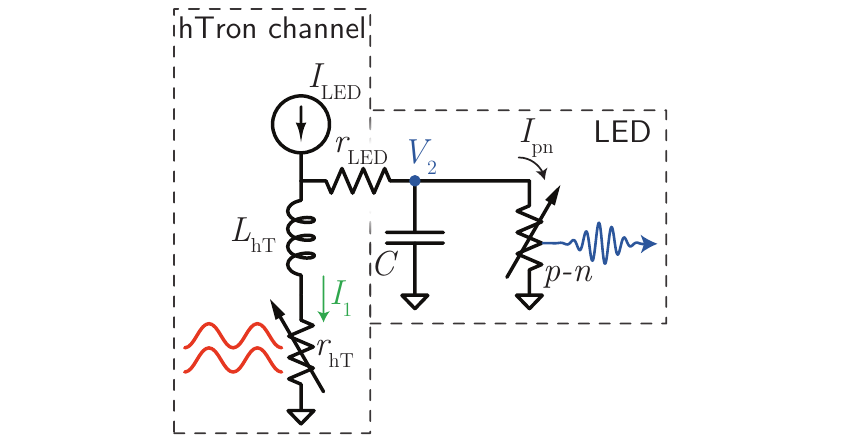}}
	\caption{\label{fig:transmitters_LED_circuit}Circuit diagram of the LED driven by the hTron. For this analysis, the hTron channel is modeled as a variable resistor ($r_{\mathrm{hT}}$) in series with an inductor ($L_{\mathrm{hT}}$). The LED is modeled as a capacitor, $C$, in parallel with a variable resistor (labeled $p-n$). The hTron variable resistor switches under the influence of heat produced by Joule heating in the gate resistive element, discussed in Sec.\,\ref{sec:hTron} (see also Appendix \ref{apx:hTronCircuitModel}). The LED variable resistor models the DC current-voltage characteristic of the $p-n$ junction (see Appendix \ref{apx:LEDCircuitModel}). The hTron channel is biased with $I_{\mathrm{LED}}$. When the channel is driven normal, $r_{\mathrm{hT}}$ switches from 0\,$\Omega$ to $800$\,k$\Omega$. The current charges up the capacitor until the voltage is $\approx$\,1\,V, at which point current begins to flow through the $p-n$ junction, producing light.}
\end{figure}

The circuit under consideration is shown in Fig.\,\ref{fig:transmitters_LED_circuit}. The LED is modeled as a variable resistor in parallel with a capacitor. The variable resistor is modeled with the DC $I-V$ characteristic of a $p-n$ junction, as described in Appendix \ref{apx:LEDCircuitModel}. The hTron is modeled as a variable resistor in series with an inductor. This variable resistor is treated as a step function with zero resistance abruptly switching to 800\,k$\Omega$. This model is intended to capture the behavior of the hTron driving the LED as the superconducting channel of the hTron is driven above its transition temperature. A thermal model of the hTron is presented in Sec.\,\ref{sec:hTron}.

The equations of motion for the circuit of Fig.\,\ref{fig:transmitters_LED_circuit} are given in Appendix \ref{apx:LEDCircuitModel}. Solving these equations, we obtain the circuit currents and voltages as a function of time. From these quantities, we can determine the energy dissipated during a firing event as well as the number of photons produced. The number of photons produced is calculated as 
\begin{equation}
\label{eq:Nph}
N_{\mathrm{ph}} = \frac{\eta_{\mathrm{qe}}}{e}\int I_{pn} dt, 
\end{equation}
where $\eta_{\mathrm{qe}}$ is the quantum efficiency of the diode, and $e$ is the electron charge. The efficiency of the circuit of Fig.\,\ref{fig:transmitters_LED_circuit} is calculated as
\begin{equation}
\label{eq:eta_RC}
\eta_{RC} = h \nu N_{\mathrm{ph}} / E_{RC}, 
\end{equation}
where $h$ is Planck's constant, $\nu$ is the frequency of a photon (taken to be $c/1.22$\,\textmu m \cite{buch2017}), and $E_{RC}$ is the total energy dissipated by the $RC$ circuit of Fig.\,\ref{fig:transmitters_LED_circuit} during a firing event as calculated with the model of Appendix \ref{apx:LEDCircuitModel}. 

At least two other factors will contribute to the efficiency of the circuit in Fig.\,\ref{fig:transmitters_LED_circuit}. First, carriers injected into the $p-n$ junction may recombine non-radiatively. This loss mechanism is captured in the internal quantum efficiency, $\eta_{\mathrm{qe}}$. Second, light generated by electron-hole recombination events may not couple to guided mode \cite{buch2017} of the axonal waveguide \cite{sh2018a,sh2018e}, and will therefore not couple to the synaptic terminals. This loss mechanism is captured in the waveguide coupling efficiency, $\eta_{\mathrm{wg}}$. The total LED efficiency is given by $1/\eta_{\mathrm{LED}} = 1/\eta_{RC}+1/\eta_{\mathrm{qe}}+1/\eta_{\mathrm{wg}}$.

\begin{figure} %[t] %[htb]
	\centerline{\includegraphics[width=8.6cm]{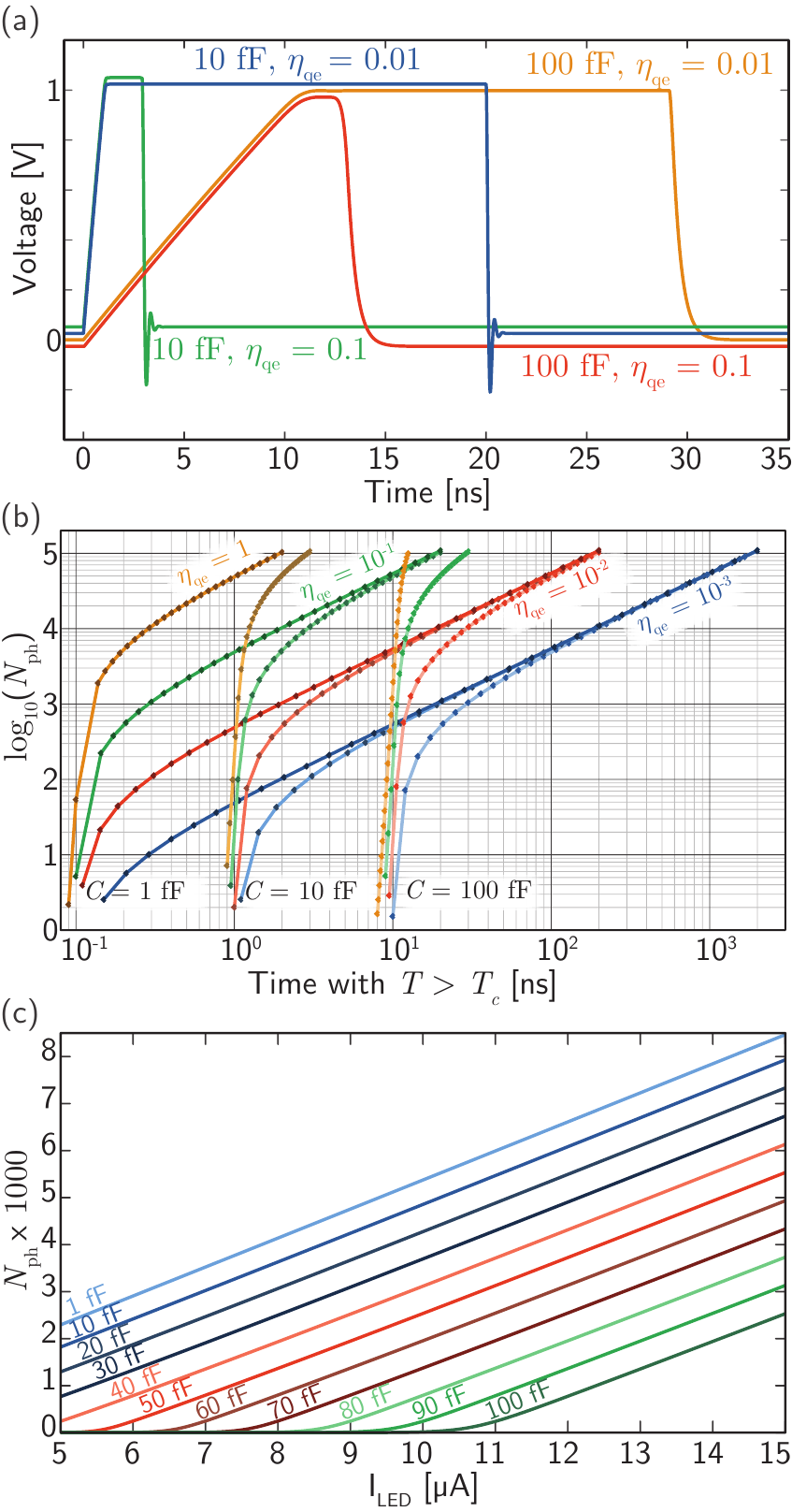}}
	\caption{\label{fig:transmitters_LED_data}Time-domain analysis of the circuit of Fig. \ref{fig:transmitters_LED_circuit}. (a) Voltage transients for two values of LED capacitance (10 fF and 100 fF) and two values of LED quantum efficiency (0.1 and 0.01). For each of the four calculations, the hTron channel resistance was modeled as a square pulse of duration necessary to achieve 10,000 photons. Traces slightly shifted in $y$ for disambiguation. (b) The number of photons produced by the LED as a function of the time the hTron channel was held above $T_c$. For short pulse durations, the traces group based on the capacitance, while for long pulse durations the traces group based on LED quantum efficiency. (c) The number of photons produced by the LED as a function of the current bias for several values of the LED capacitance. Here the hTron channel was held above $T_c$ for 10\,ns. The slope of the traces ranges from 617 photons per \textmu A with $C = 1$\,fF to 584 photons per \textmu A with $C = 100$\,fF. }
\end{figure}

Circuit performance is shown in Fig.\,\ref{fig:transmitters_LED_data}. Voltage transients across the $p-n$ junction are shown in Fig. \ref{fig:transmitters_LED_data}(a), where Two values of capacitance and two values of quantum efficiency are considered. For each of the four traces, the hTron pulse duration was chosen to produce 10,000 photons. In this paper and the next in this series \cite{sh2018e}, we assume that 10 photons are generated per out-directed synaptic connection to compensate for loss, to reduce noise, and to implement learning functions. Thus, voltage pulses such as these are appropriate to neurons with out-degree of 1000. If the LED capacitance can be made as low as 10\,fF, and $\eta_{\mathrm{qe}}$ can be made as high as 0.1, the hTron gate must only be normal for 2.9\,ns. Under these operating conditions, 25\,fJ of energy is dissipated, of which 64\% is dissipated as current through the $p-n$ junction ($\eta_{RC}=0.64$). If the LED performance is worse, with $C = 100$\,fF and $\eta_{\mathrm{qe}} = 0.01$, the drive must be applied for 29\,ns. This operation dissipates 251\,fJ of energy, also with 64\% through the junction.

General trends for the number of photons produced by the circuit as a function of the time the hTron gate is above $T_c$ are shown in Fig.\,\ref{fig:transmitters_LED_data}(b) for several values of capacitance and quantum efficiency. In these calculations, the bias current ($I_{\mathrm{LED}}$) is fixed at 10\,\textmu A. The capacitance determines the minimum hTron pulse duration necessary to achieve the voltage necessary for photon production, and the quantum efficiency determines the slope ($y$-intercept of this log-log plot) for longer hTron pulses.

As implied by the gain drive current depicted in Fig.\,\ref{fig:transmitters_schematic}(a), we would like a means to dynamically control the number of photons produced in a firing event. We can achieve this by varying the bias current diverted to the LED when the channel of the hTron becomes resistive, $I_{\mathrm{LED}}$. In Fig.\,\ref{fig:transmitters_LED_data}(c) we show $N_{\mathrm{ph}}$ as a function of $I_{\mathrm{LED}}$ for various values of the LED capacitance in a case where the hTron channel is driven normal for 10\,ns and the LED quantum efficiency is assumed to be 0.01 \cite{doro2017}. This current bias provides a means to change the strength of neuronal activity. With fewer photons produced, the probability of reaching distant connections diminishes. With more photons produced, the neuron makes a stronger contribution to network activity. This current bias may be modified by a supervisory user or by internal activity within the network.

The efficiency of the light-production circuit is considered with the rest of the circuit in Sec.\,\ref{sec:efficiency}. Based on the calculations of this section, we know the hTron drive requirements for light production across a range of capacitance and quantum efficiency values. If a neuron needs to produce 10,000 photons to communicate to 1000 synaptic connections, an hTron biased with a channel current of 10\,\textmu A must produce a resistance of 800\,k$\Omega$ for 1\,ns - 100\,ns, depending on the achievable LED performance. We now proceed to consider the operation of the hTron when driven by the current from an nTron.  
	
\section{\label{sec:hTron}The hTron voltage amplifier driven by the nTron current amplifier}
\begin{figure} %[t] %[htb]
	\centerline{\includegraphics[width=8.6cm]{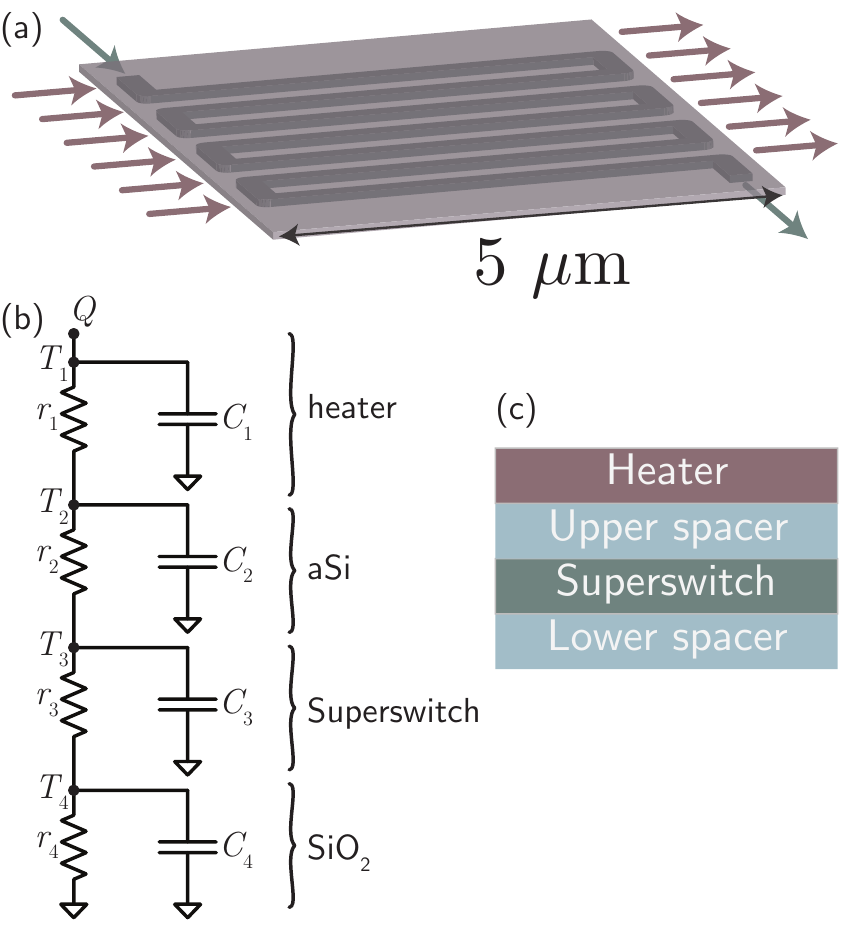}}
	\caption{\label{fig:transmitters_hTron_circuit}The hTron voltage amplifier. (a) Schematic of the thin-film configuration with a normal-metal resistive layer (gate) over the superconducting nanowire meander (channel). (b) Thermal circuit model used to calculate the dynamics of the device. (c) Cross-sectional view of layers involved in calculation.}
\end{figure}

As discussed in Sec.\,\ref{sec:LED}, in order to produce the voltage across the LED required to generate light, a resistance of 800\,k$\Omega$ must be established rapidly and sustained for 1\,ns - 100\,ns. One way to achieve this is to switch a length of superconducting wire to the normal metal state. This can be straightforwardly accomplished by raising the temperature of a superconducting wire above $T_c$. An hTron is a device that switches a channel resistance from zero ohms to a large value when current is driven through the gate \cite{zhto2018}. The current through the gate dissipates power in a resistive element through Joule heating, and this power locally raises the channel above $T_c$. We show a schematic in Fig.\,\ref{fig:transmitters_hTron_circuit}(a), wherein a resistive layer is separated from a nanowire meander by a thin insulator. 

While thermal devices can be slow and inefficient in some contexts, there are several reasons why the hTron is suitable for the present purpose. First, the device has a compact footprint of 5\,\textmu m $\times 5$\,\textmu m (see Appendix \ref{apx:hTronCircuitModel}), so a very small mass must be heated. Second, the specific heat of all materials involved falls as $T^3$, so at the desired operating temperature of 4.2\,K, the specific heat is orders of magnitude smaller than at room temperature. Third, the required temperature swing is small ($\approx 2$\,K), so little power dissipation is required to raise the device above $T_c$. These factors taken together make the hTron a suitable device to achive the voltage necessary to produce light from a semiconductor diode with the power and speed required for the neural application under consideration.

To quantify the performance of the hTron when driving the LED of Sec.\,\ref{sec:LED}, we consider the transient dynamics of the thermal circuit shown in Fig.\,\ref{fig:transmitters_hTron_circuit}(b). A heat source $Q$ is delivered to the stack of materials shown in Fig.\,\ref{fig:transmitters_hTron_circuit}(c). The equations of motion and material parameters used in these calculations are given in Appendix \ref{apx:hTronCircuitModel}. Figure\,\ref{fig:transmitters_hTron_data_1}(a) shows temperature transients of the superconducting layer when power is delivered to the hTron gate long enough to drive the hTron channel normal for 1\,ns and for 10\,ns. In this plot, the hTron gate is driven with square current pulses to illustrate the temporal dynamics of the thermal components. With this model, we see the hTron can switch to the resistive state in roughly 1\,ns. While this time scale is not suitable for many operations in superconducting digital electronics, it is more that fast enough for a neuronal firing event in the system under consideration.  
\begin{figure} %[t] %[htb]
	\centerline{\includegraphics[width=8.6cm]{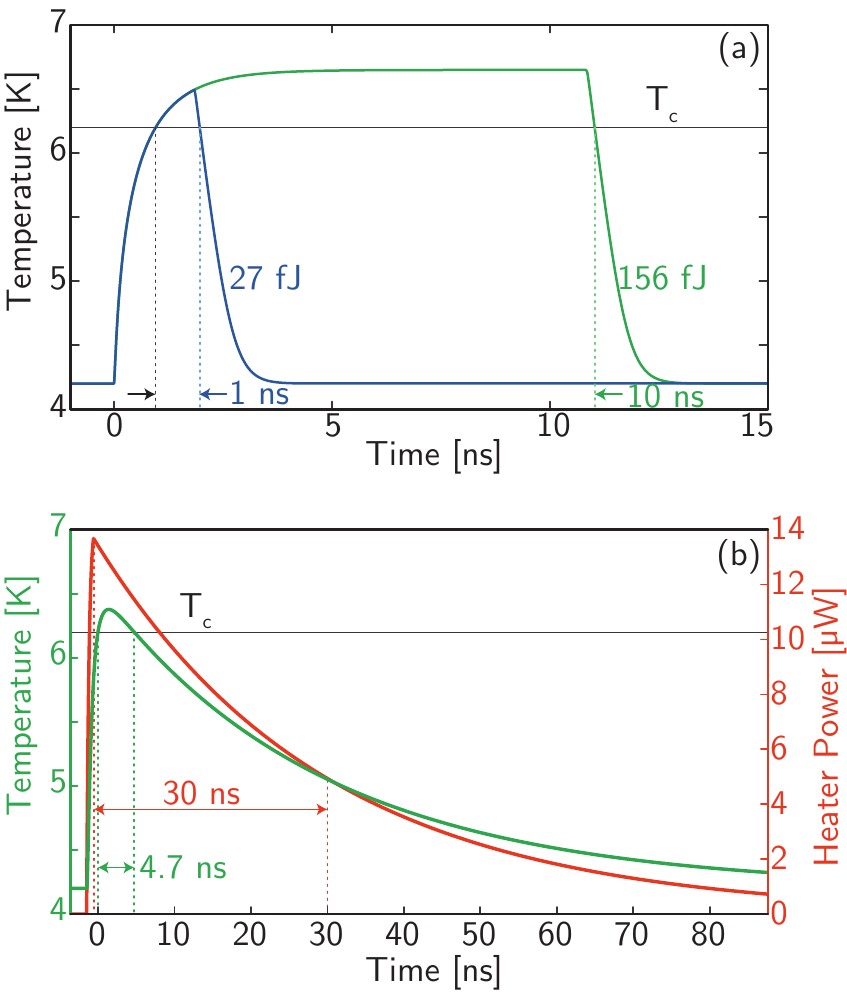}}
	\caption{\label{fig:transmitters_hTron_data_1}Time-domain analysis of the hTron. (a) A square current pulse is driven through the gate of the hTron. Two traces are shown for cases when the channel was driven normal long enough to keep the the temperature above $T_c$ for 1\,ns and 10\,ns. (b) An exponential current pulse is driven through the gate of the hTron, as would be the case when the nTron is used to drive the hTron. In the case shown, the $L/r$ time constant of the nTron is 30\,ns, and the channel of the hTron is held above $T_c$ for 4.7\,ns.}
\end{figure}

While a square pulse is most efficient for driving the hTron, exponentially rising and decaying pulses are more straightforward to achieve. The nTron current amplifier we plan to employ to produce such pulses was introduced in Ref.\,\onlinecite{mcbe2014}. The nTron is a three-terminal, thin-film device. In the off state, a supercurrent flows from the source to drain, and the gate is in the superconducting state. The gate comprises a small constriction, and when the current delivered to the gate exceeds the critical current, it is driven locally to the normal state, and Joule heating spreads the normal domain. This quickly causes the channel between the source and drain to be driven normal across the entire wire width, leading to a few squares of resistance (roughly 1\,k$\Omega$), at which point the source-drain current is diverted to a load. In the present case, the load is the gate of the hTron (10\,$\Omega$ in this study, the resistance of the upper layer in Fig.\,\ref{fig:transmitters_hTron_circuit}(a) and (c)). When the gate current to the nTron ceases and the channel returns to the superconducting state, the channel current returns with the $L/r$ time constant of the system. Thus, the current pulse from the nTron to the hTron will have an exponential time dependence. 

In Fig.\,\ref{fig:transmitters_hTron_data_1}(b) we show the temporal response of the hTron channel temperature when the gate is driven by an exponential current pulse of 1.2\,mA amplitude from an nTron. In this case, the rise time of the pulse is 300\,ps, and the fall time is 30\,ns. These time constants are controlled by the $L/r$ time constants of the circuit. The fall time is set by $L_{\mathrm{nT}}/r_{\mathrm{nT}}$, where $L = L_{\mathrm{nT}}$ the inductance of the nTron channel, and $r_{\mathrm{nT}}$ is the load of the nTron, which in this case is the gate resistance of the hTron. To achieve the power necessary to switch the hTron, 1.2\,mA from the nTron is required across the 10\,$\Omega$ of the hTron gate.

\begin{figure} %[t] %[htb]
	\centerline{\includegraphics[width=8.6cm]{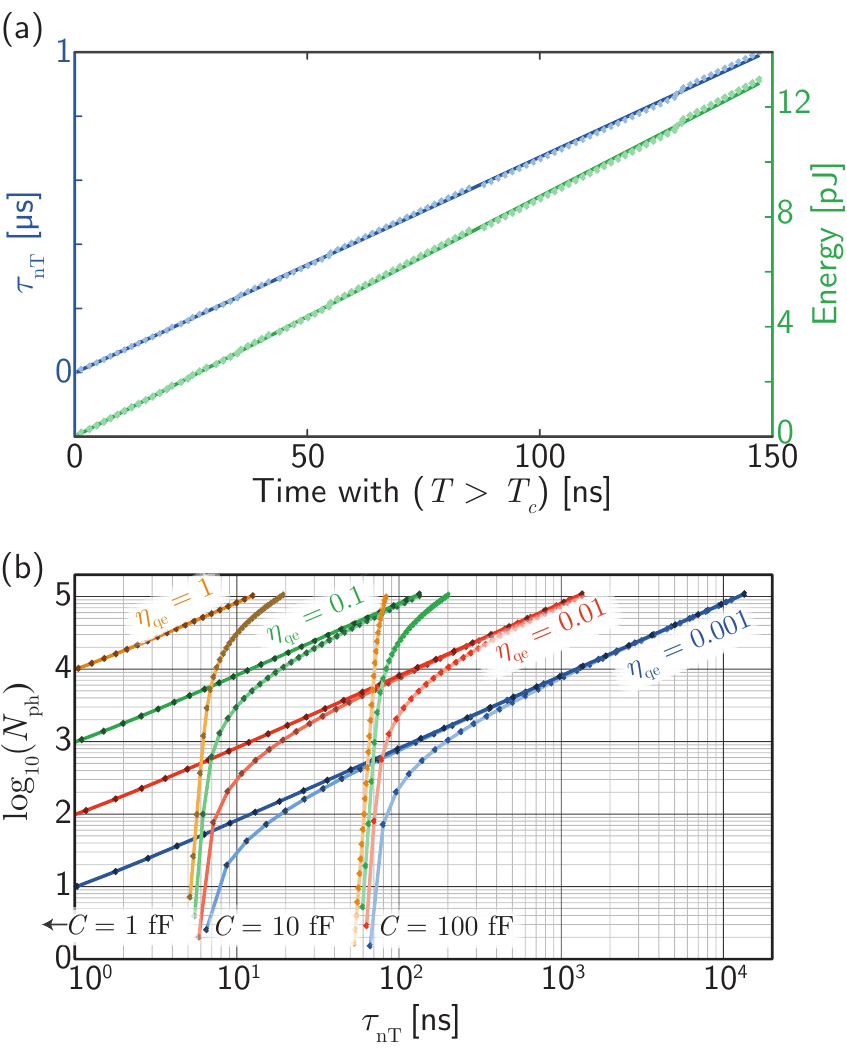}}
	\caption{\label{fig:transmitters_hTron_data_2}Energy analysis of the hTron. (a) nTron time constant required and energy consumed as a function of the time the hTron is held above $T_c$. Dots are data from discrete numerical calculations, and solid lines are linear fits. The energy efficiency is nearly an order of magnitude higher when the hTron is driven by a square pulse. (b) The number of photons produced by the LED as a function of the nTron $L/r$ time constant, $\tau_{\mathrm{nT}}$, for several values of LED efficiency and capacitance.}
\end{figure}

Driving the hTron with the exponential pulses of the nTron is far less efficient than driving with square pulses because power continues to be dissipated in the exponential tail of the current pulse long after the temperature of the hTron channel drops back below $T_c$. In the case considered in Fig.\,\ref{fig:transmitters_hTron_data_1}(b), the $L_{\mathrm{nT}}/r_{\mathrm{nT}}$ time constant is 30\,ns, and the hTron channel is held above $T_c$ for 4.7\,ns. A square pulse of 6\,ns would achieve the same duration of the hTron in the resistive state. Figure \ref{fig:transmitters_hTron_data_2}(a) illustrates this inefficiency. The required $L_{\mathrm{nT}}/r_{\mathrm{nT}}$ time constant is plotted as a function of the time the hTron channel must be held above $T_c$, referred to as $t_{>}$. The time $t_{>}$ depends on the LED capacitance, efficiency, and the number of photons produced during the pulse, which is determined by the number of synapses formed by the neuron. The $L_{\mathrm{nT}}/r_{\mathrm{nT}}$ time constant must be nearly ten times $t_>$. Improved drive circuit designs are likely possible, but we proceed with the single-nTron example to illustrate that even with first-generation circuit designs, sufficient efficiency can be achieved. 

We wish to connect the number of photons produced in a neuronal firing event to the $\tau_{\mathrm{nT}}$ time constant that must be implemented in hardware. The number of photons required depends on the number of synaptic connections as well as the LED capacitance and efficiency. In Ref.\,\onlinecite{sh2018e} we study networks with neurons making 20-1000 synaptic connections. We would like neuronal firing events to produce $10\times$ as many photons as the neuron's out-degree to compensate for loss, reduce noise, and perform memory update operations. We are therefore interested primarily in neuronal firing events producing 200-10,000 photons, and large hub neurons may need to produce 100,000 photons or more. Such photon numbers will require the hTron channel to be driven normal for some duration, which necessitates current drive from the nTron to the hTron gate for some proportional duration (Fig.\,\ref{fig:transmitters_hTron_data_2}(a)). We therefore must calculate the number of photons produced by the LED as a function of the $\tau_{\mathrm{nT}} = L_{\mathrm{nT}}/r_{\mathrm{nT}}$ time constant, taking the LED capacitance and efficiency into account. The results of these calculations are shown in Fig.\,\ref{fig:transmitters_hTron_data_2}(b). To produce 200 photons from an LED with 10\,fF capacitance and 1\% efficiency, $\tau_{\mathrm{nT}}$ must be 9\,ns, and to produce 10,000 photons $\tau_{\mathrm{nT}}$ must be 100 ns. We show in Appendix \ref{apx:hTronInductor} that an nTron channel inductor achieving this time constant and carrying th requisite 1.2\,mA necessary to drive the hTron gate can be achieved in an area commensurate with the area required for the synaptic wiring and routing waveguides comprising the dendritic and axonal arbors of the loop neurons. 

At this point we have computationally demonstrated an nTron driving an hTron driving an LED. Assessing the contributions of the nTron and the hTron to the total energy of a neuronal firing event must occur in the context of the firing circuit as a whole. We consider the various contributions to neuronal firing energy and photon production efficiency in Sec.\,\ref{sec:efficiency}. The next task is to show a Josephson junction in the neuronal thresholding loop delivering the current to switch the gate of the nTron.
	
\section{\label{sec:detectingThreshold}Detecting neuronal threshold}
The circuit considered for detecting threshold in the neuronal thresholding loop is shown in Fig.\,\ref{fig:transmitters_amplifierChain_circuit}. The parameters are given in Appendix \ref{apx:roParams}. When the current in the neuronal integrating (NI) loop reaches the switching current of the thresholding junction, $J_{\mathrm{th}}$, that junction will switch and transmit a fluxon to the relaxation oscillator junction, $J_{\mathrm{ro}}$. The relaxation oscillator junction is designed to enter a transient resistive period upon arrival of a fluxon from $J_{\mathrm{th}}$. The current from $J_{\mathrm{ro}}$ switches the gate of the nTron, diverting the nTron channel current to the gate of the hTron. 

\begin{figure} %[t] %[htb]
	\centerline{\includegraphics[width=8.6cm]{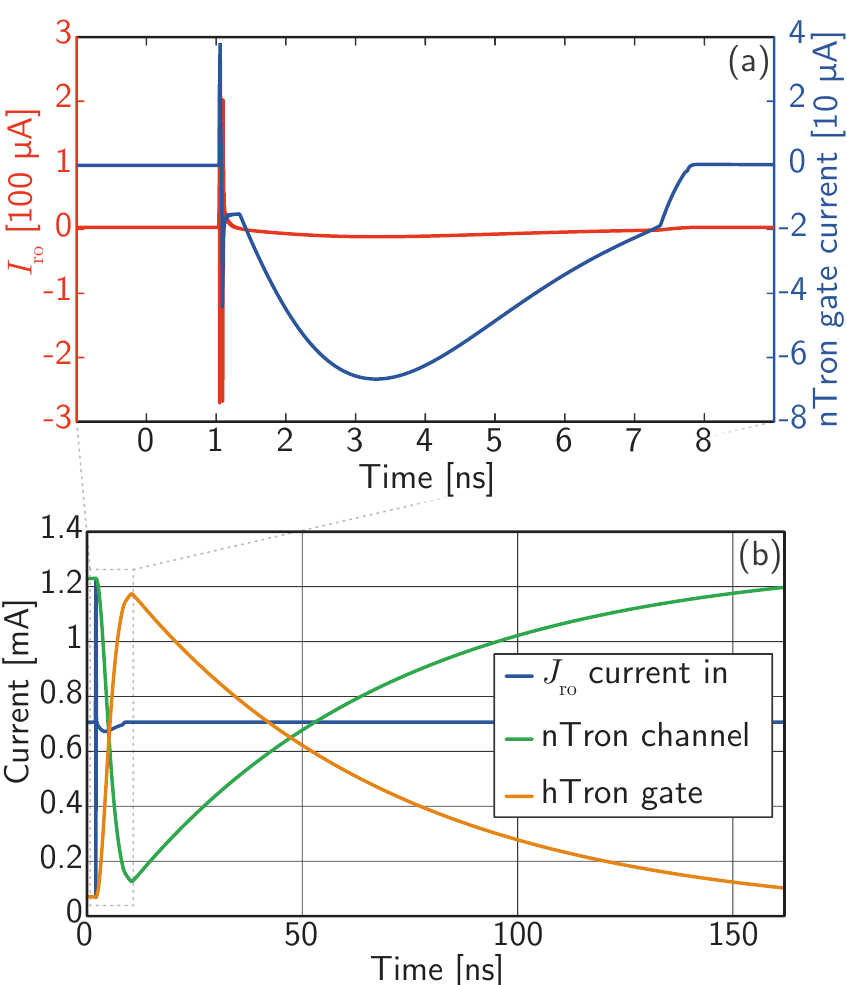}}
	\caption{\label{fig:transmitters_jjTriggersNTron}Time-domain simulations of the threshold detection event. (a) A fluxon from the threshold junction $J_{\mathrm{th}}$ switches $J_{\mathrm{ro}}$. The current diverted from $J_{\mathrm{ro}}$ to the gate of the nTron is plotted as a function of time. (b) The nTron source/drain current and the hTron gate current as a function of time following the threshold detection event. The simulation shown in Fig.\,\ref{fig:transmitters_jjTriggersNTron} was initiated with a DC pulse of 50\,mV into a DC-to-SFQ converter \cite{vatu1998,ka1999} that produced a fluxon. This fluxon propagated through a Josephson transmission line (JTL) \cite{vatu1998,ka1999} before switching the voltage-state junction. This JTL is not essential for operation, and it adds negligible power consumption to the threshold detection event.}
\end{figure}

Spice simulations of the thresholding event are shown in Fig.\,\ref{fig:transmitters_jjTriggersNTron}. For these simulations, Cadence was used. In Fig.\,\ref{fig:transmitters_jjTriggersNTron}(a), the thresholding junction switches, producing a current pulse from the relaxation oscillator junction. In Fig.\,\ref{fig:transmitters_jjTriggersNTron}(b), the gate of the nTron switches, and the nTron channel current is diverted to the gate of the hTron, returning with the $\tau_{\mathrm{nT}}$ time constant. 

The series of events shown in Fig.\,\ref{fig:transmitters_jjTriggersNTron} comprise the detection of neuronal threshold, which occurs when $J_{\mathrm{th}}$ switches, and the subsequent current amplification, beginning with the switching of $J_{\mathrm{ro}}$, and leading to the switching of the nTron. The 1.2\,mA current pulse coming from the nTron and driving the gate of the hTron has been shown in Sec.\,\ref{sec:hTron} to be sufficient to switch the hTron, resulting in a voltage pulse and light generation from the LED. The $L/r$ time constant in this simulation was 50\,ns, corresponding to 500 nH in series with the 10\,$\Omega$ of the hTron gate. Figure\,\ref{fig:transmitters_hTron_data_2}(b) shows this nTron time constant is sufficient to produce more than 3000 photons if the LED capacitance is 10\,fF and the LED efficiency is 1\%. Increasing the nTron inductance can extend the pulse duration and thereby produce more photons. With the material and device parameters used in the model of this nTron, 50\,ns recovery was sufficient to keep the nTron from latching, and shorter recovery times may be possible, depending on electro-thermal device engineering. As discussed in Sec.\,\ref{sec:hTron}, shaping the nTron pulse to be square rather than exponential would have a significant impact on the light-production efficiency. Such operation may be achieved with circuits utilizing two or more nTrons with feedback.

Having demonstrated threshold detection, current amplification, and voltage amplification, we have completed the description of the amplification chain of the neuronal transmitter circuit. We next discuss the efficiency of the amplifier chain.
	
\section{\label{sec:efficiency}Photon production efficiency}
In Secs.\,\ref{sec:LED}-\ref{sec:detectingThreshold} we have described circuits capable of detecting threshold in the neuronal threshold loop, amplifying the signal, and producing the voltage necessary to generate light from a semiconductor diode. We have calculated the energy consumed by each element of the amplifier chain when generating the number of photons necessary for neuronal communication in realistic networks. The networks we wish to consider are described in Ref.\,\onlinecite{sh2018e}. As explained in Ref.\,\onlinecite{sh2018a}, they will comprise a variety of neurons with a range of numbers of synapses. The number of out-directed synapses made by a neuron is referred to as the out-degree, $k_{\mathrm{out}}$. The number of photons that must be produced in a neuronal firing event depends on $k_{\mathrm{out}}$, and therefore so does the energy of a neuronal firing event, which we denote by $E_{\mathrm{out}}(k_{\mathrm{out}})$.

\begin{figure} %[t] %[htb]
	\centerline{\includegraphics[width=8.6cm]{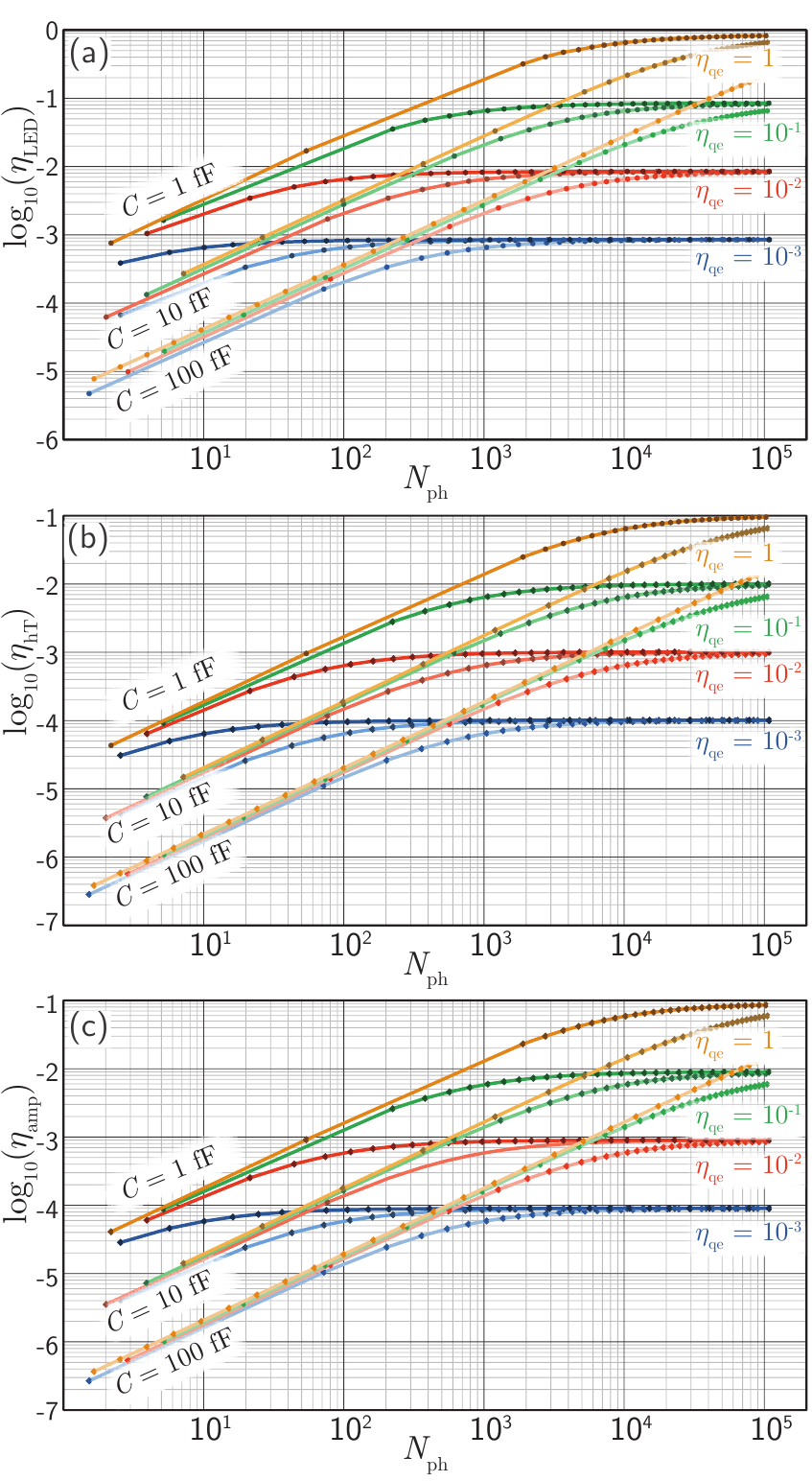}}
	\caption{\label{fig:transmitters_efficiency}Efficiency of the photon production process as a function of the number of photons produced. (a) Efficiency of the LED. (b) Efficiency of the hTron. (c) Combined efficiency of the LED and hTron.}
\end{figure}
In an ideal case, a neuron could produce one photon per synaptic connection with unity production efficiency, the photons would reach their destination without loss, and they would be detected at the synapse with unity detection efficiency. In this case, we would have $E_{\mathrm{out}} = h\nu k_{\mathrm{out}}$, where $h$ is Planck's constant and $\nu$ is the frequency of the photons being generated. In an actual network implemented in hardware, waveguides will have propagation loss, and detectors will have efficiency less than unity. To account for these loss mechanisms, a neuron will produce a number of photons greater than one per synaptic connection. We refer to this number of photons as $\zeta$. If $\zeta$ is too large, neurons are wasting power. If $\zeta$ is too small, communication will be unreliable. The synapses of Ref.\,\onlinecite{sh2018b} have the same response if they receive one or more photons. Therefore, the noise is not shot noise, as it would be if the synapse were attempting to detect the precise number of incident photons. The communication error can be calculated from the Poisson distribution. Given an average number of incident photons on a synapse due to an upstream neuronal firing event, we use the Poisson distribution to calculate the probability that a synapse will receive zero photons due to a neuronal firing event. If the average number of incident photons is five, the probability that the synapse will receive zero photons is less than 1\%. A neural system is likely to be able to tolerate this level of error \cite{stgo2005}. In the case of lossy waveguides with low-efficiency detectors, 3\,dB loss may be incurred between a neuron and a synaptic target. We thus consider $\zeta = 10$ to be a conservative number to use in calculations of network power consumption.

In addition to propagation loss and detector inefficiency, the circuits described in this work are not entirely efficient at producing photons. If an LED produces $N_{\mathrm{ph}}$ photons during a firing event, and the LED consumes $E_{\mathrm{LED}}$ total energy during that firing event, the LED efficiency is defined by $E_{\mathrm{LED}}(N_{\mathrm{ph}}) = h \nu N_{\mathrm{ph}}/\eta_{\mathrm{LED}}(N_{\mathrm{ph}})$.  A similar expression holds for the nTron driving the hTron. The total energy consumed by the amplifier chain during a neuronal firing event is given by
\begin{equation}
\label{eq:energy}
E_{\mathrm{amp}}(N_{\mathrm{ph}}) = \zeta h \nu N_{\mathrm{ph}}/\eta_{\mathrm{amp}}(N_{\mathrm{ph}}),
\end{equation}
where $1/\eta_{\mathrm{amp}} = 1/\eta_{\mathrm{LED}}+1/\eta_{\mathrm{hT}}$. The Josephson circuits driving the nTron contribute negligible energy consumption. The calculations of Secs.\,\ref{sec:LED} and \ref{sec:hTron} provide us with numbers for $\eta_{\mathrm{LED}}$ and $\eta_{\mathrm{hT}}$ for several values of LED capacitance and internal quantum efficiency. These functions are plotted in Fig.\,\ref{fig:transmitters_efficiency}(a) and (b). Figure\,\ref{fig:transmitters_efficiency}(c) shows the total amplifier efficiency, $\eta_{\mathrm{amp}}$. We see that driving the hTron with the nTron (Fig.\,\ref{fig:transmitters_efficiency}(b)) dominates power consumption, yet the amount of time the hTron must be on is determined by the emitter capacitance and efficiency, and therefore emitter improvement is necessary for system improvement. The circuits can also be more efficient with improved thermal design of the hTron (through phonon localization) and current pulse shaping from the nTron.

In practice, achieving small, waveguide-integrated LEDs with low capacitance should be possible. A simple parallel plate model indicates 1\,fF should be achievable, and 10\,fF should not be particularly challenging, even with wiring parasitics. In Fig.\,\ref{fig:transmitters_efficiency}, we consider values as poor as 100\,fF. The quantum efficiency of the device is harder to predict. Waveguide-integrated light-emitting diodes with efficiency near 0.01 have been demonstrated \cite{doro2017}, and low-temperature operation helps significantly in this regard. We expect quantum efficiency of 0.01 to be achievable at large scale, but in Fig.\,\ref{fig:transmitters_efficiency} we consider values as poor as $10^{-3}$. Considering an LED with 10 fF capacitance or less, if the LED efficiency is as poor $10^{-3}$, the total photon production efficiency of the amplifier chain is $\eta_{\mathrm{amp}} \approx 10^{-4}$ when more than 200 photons are produced. For calculating the power consumption of the networks described in Ref.\,\onlinecite{sh2018e}, we use Eq.\,\ref{eq:energy} with $\zeta = 10$ and $\eta_{\mathrm{amp}} = 10^{-4}$. We next discuss reset of the device and the inclusion of a refractory period.
		
\section{\label{sec:refraction}Reset and refraction}
\begin{figure} %[t] %[htb]
	\centerline{\includegraphics[width=8.6cm]{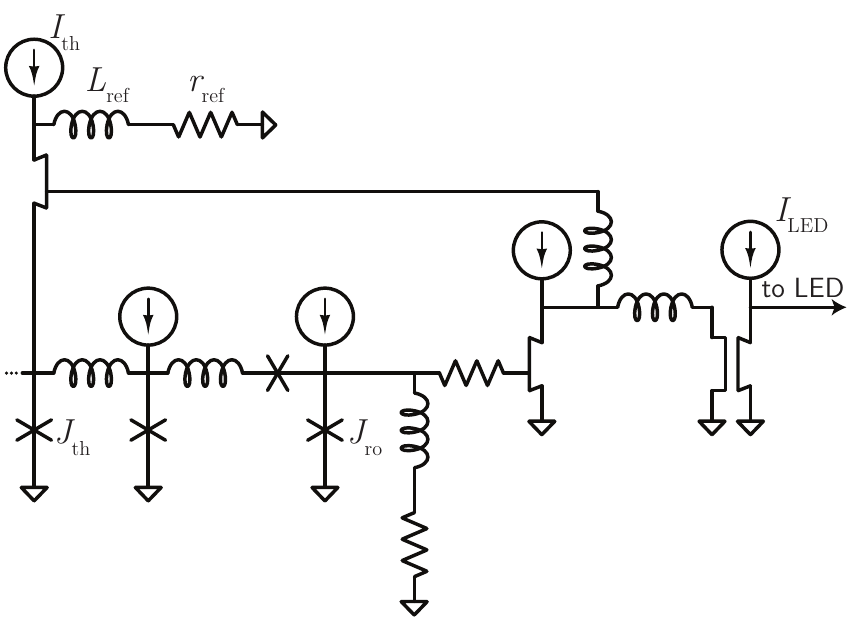}}
	\caption{\label{fig:transmitters_threshold_circuit}Variant on the circuit in Fig.\,\ref{fig:transmitters_amplifierChain_circuit}. Here a refractory period is implemented by cutting off current to $J_{\mathrm{th}}$ with an additional nTron. Parameters used in the model are given in Appendix \ref{apx:roParams}.}
\end{figure}

We have described the production of light by the transmitter circuit, and we must now consider how the circuit resets. Reset is an important part of a neuron's function, and in the integrate-and-fire model, it is often assumed that upon reaching threshold, the variable representing the integrated signal is instantaneously set to some initial value \cite{daab2001,geki2002}. In the circuit of Fig.\,\ref{fig:transmitters_threshold_circuit}, the switching of $J_{\mathrm{th}}$ results in the addition of flux to the NI loop, countering the applied current bias. Such operation may be useful for reset. 

Yet it behooves us to consider reset operation in the broader context of continuous afferent synaptic activity to the loop neuron. Assume the synapses described in Ref.\,\onlinecite{sh2018b} and connected to the NI loop receive inputs at a constant rate, and the synaptic integration (SI) loops have a constant leak rate, set by an $L/r$ time constant. At any moment in time, we would like the neuronal thresholding loop to contain a current that is a linear combination of the currents in the SI loops (or the dendritic integrating loops if they are employed \cite{sh2018b}). Now suppose all afferent stimuli cease, and the currents in the SI loops decay to zero. If flux has been trapped in the NI loop due to a neuronal thresholding event, $J_{\mathrm{th}}$ will be biased below its original bias point by the current of a flux quantum, $\Phi_0/L_{\mathrm{ni}}$. It may be advantageous to employ a buffer junction inside the NI loop to switch in response to the fluxon from $J_{\mathrm{th}}$, thereby leaving the NI loop in the same state as immediately before the thresholding event. 

In this manner of operation, the SI and NT loops always maintain a representation of the state of the synapses. In order to ensure spiking operation and keep the circuit from generating a constant stream of photons when threshold is reached, refraction can be introduced and controlled with the circuit shown in Fig.\,\ref{fig:transmitters_threshold_circuit}. When threshold is reached and the nTron switches, some of the current is diverted to a choke nTron above the thresholding junction. The time constant, $L_{\mathrm{ref}}/r_{\mathrm{ref}}$, sets the refractory period.

Another means to detect threshold while avoiding adding flux to the neuronal integrating loop is to use a yTron \cite{mcab2016} as the thresholding element instead of the thresholding JJ. The yTron has the advantage of providing the non-destructive readout we seek, and a yTron could likely drive the gate of the nTron without the need for a latching junction. The major challenge of the yTron is that if it is going to be induced to threshold with a small current, as appears to be beneficial from the consideration of multisynaptic neurons in Ref.\,\onlinecite{sh2018b}, the yTron gate will require small features and a sharp corner \cite{mcab2016} as well as low inductance. A DC SQUID \cite{ti1996,vatu1998,ka1999} is also a candidate for non-destructive readout of the NT loop. In this variant, the NT loop would be a DC SQUID, and the voltage across it would represent the integrated signal from the SI loops. When sufficient voltage was reached, a junction in parallel would switch, leading to the amplification sequence described in this paper.

At present it appears that several approaches to threshold, reset, and refraction are viable based on circuits with Josephson junctions and thin-film amplifiers. This discussion has considered readout and reset of the NT loop, but we must also consider the behavior of the SI loops after neuronal firing. Many models exist \cite{ma1997,geki2002} for treating synaptic dynamics, and it may be advantageous to utilize different synapses within a system or within a neuron with different reset dynamics. In the present context of loop neurons wherein afferent synaptic activity is integrated in an ensemble of SI loops, we assume each synapse has an independent leak rate set by the $L/r$ time constant of that loop. We may think of an SI loop as a device providing information regarding its recent history of activity within a sliding time window set by the time constant. This information represents not just the integrated signal, but also may include higher-order temporal terms obtained through coincidence detection \cite{sh2018b}, dendritic processing \cite{sh2018b}, and short-term plasticity \cite{sh2018c}. The dendritic arbor, comprising all the neuron's synapses, contains information about not just recent average activity, but also higher-order temporal statistics. Neuronal refraction is necessary to ensure spiking activity and not latching behavior, but if the circuits do not reset the synapses after each neuronal firing event, the dendritic arbor maintains a sliding history of synaptic activity unperturbed by neuronal firing events. The dendritic arbor is a spatial distribution of synapses that contain a broad representation of information from the frequency domain. The NI loop and NT loop preserve mirrors of the net dendritic flux. In the NT loop, this requires a buffer to release flux from a neuronal firing event. The threshold, determined by the difference between the critical current of the threshold junction and bias to the threshold junction, experiences first an absolute then a relative refractory period as current returns to the threshold junction with an $L/r$ time constant.

The leak rates of the synaptic integrating loops and the refractory period of the neuron set the time constants that determine the nonlinearities in the rate in/rate out transfer function of the system and contribute to the oscillation frequencies of the relaxation oscillators. The leak rates of the integrating loops can be tuned from DC to MHz, enabling the turn-on nonlinearity of the rate in/rate out neuronal response to be placed across a wide range of frequencies. The refractory period can be lengthened with the $L_{\mathrm{ref}}/r_{\mathrm{ref}}$ time constant mentioned above, and it can be as short as the recovery time of the nTron. The hTron is only driven normal during a brief duration of the nTron exponential pulse, and there is a refractory period associated with the current returning to the nTron channel, as the nTron gate cannot be switched if the bias to the nTron is not above a certain value. It may be possible to reduce this recovery time to 10 ns with proper materials and design, and if the LEDs can be made efficient, such short pulses may produce sufficient photons to drive the neuron's synaptic connections. In this case the maximum neuronal firing rate would be 100 MHz. Yet it remains to be demonstrated conclusively that such rapid operation is possible. The nTron investigated in Fig.\,\ref{fig:transmitters_jjTriggersNTron} operates with 50 ns recovery time. We therefore take 20 MHz to be the maximum oscillation frequency we consider for loop neurons in the power calculations presented in Ref.\,\onlinecite{sh2018e}.
	
\section{\label{sec:discussion}Discussion}
We have presented a circuit that amplifies the small current pulse produced when threshold is reached in the superconducting neuronal threshold loop \cite{sh2018b}. This circuit produces a voltage sufficient to generate light from a semiconductor diode. We have analyzed the temporal dynamics of the amplifier chain as well as the energy consumption to arrive at a system photon production efficiency. This efficiency will be used in the next paper in this series \cite{sh2018e} in the calculation of power consumption of a network.

The analysis presented here points to two conclusions. The first conclusion is that to produce photons with short pulses from the hTron, it is necessary to fabricate LEDs with low capacitance. Other integrated photonic applications find that capacitance dictates the scale at which light becomes advantageous \cite{mi2017}. The second conclusion is that the overall photon production efficiency is limited by the internal quantum efficiency of the LED when producing large numbers of photons. To be useful, superconducting optoelectronic neurons must function in a variety of contexts. As argued in Ref.\,\onlinecite{sh2018a}, cognitive neural systems are likely to utilize some neurons with relatively small out-degree making only local connections as well as other neurons with high degree making local and long-range connections. The value of capacitance achievable by the LED will determine the lowest degree practical to implement with photonic connectivity. A simple parallel-plate model of the LED (Appendix \ref{apx:LEDCircuitModel}) shows that achieving 1 fF should be possible, in which case neurons with as few as 10 out-directed connections may be practicable. Whatever this limit may be, we anticipate combining superconducting optoelectronic neurons in networks also employing purely superconducting electronic neurons \cite{hias2007,crsc2010,segu2014,ru2016,scdo2018} that are well-suited to forming dense local clusters with low-degree connectivity, high speed, and excellent energy efficiency. 

While capacitance limits utility for low-degree operation, internal quantum efficiency determines system power consumption dominated by neurons with high degree. The maximum achievable quantum efficiency will depend on emitter design, materials employed, and fabrication optimization. Design of emitters specifically for this application has been described in Ref.\,\onlinecite{bu2018}, where it is concluded that non-resonant devices with high fabrication yield may achieve above 40\% optical collection efficiency. The electrical injection efficiency will depend on many factors, and a summary of candidate emitters is presented in Ref.\,\onlinecite{shbu2017}. As in many systems, trade-offs must be weighed. Compound semiconductor light emitters are far superior to silicon light emitter in terms of efficiency, speed, and spectral coverage, yet fabrication with compound semiconductors is more expensive and more difficult to scale than silicon. While silicon has a long history as a disappointing light emitter \cite{da1989,shxu2007}, it may be good enough \cite{buch2017} for near-term development of superconducting optoelectronic networks. The ease of integration with superconducting electronics and passive photonic waveguides makes cryogenic silicon light sources legitimate as a starting point for developing this platform. To assess the long-term potential of superconducting optoelectronic networks, much more experimental work on low-capacitance, high-efficiency, waveguide-integrated few-photon light-emitting diodes is needed. Further investigation of thin-film superconducting amplifiers is also required.

The reader may wonder why we insist on incorporating a light source at each neuron. Perhaps we can utilize an architecture with circulating light in waveguides that are tapped by a modulator during each neuronal firing event. Such a system would have the benefit of keeping light sources out of the cryostat and separate from superconducting circuits. In the case of shared off-chip light sources, the new challenges become to develop very low loss waveguides; compact, efficient modulators with low insertion loss that do not need to be tuned; and extensive fiber-to-chip coupling. Initial calculations indicate that such an approach is less efficient unless waveguides can be made with very low propagation loss, and modulators can be made with very low insertion loss. Yet device and system limits are far from understood. Amplifier circuits similar to those presented here may be useful when driving the modulators of such a system.

\begin{figure*} %[t] %[htb]
	\centerline{\includegraphics[width=17.2cm]{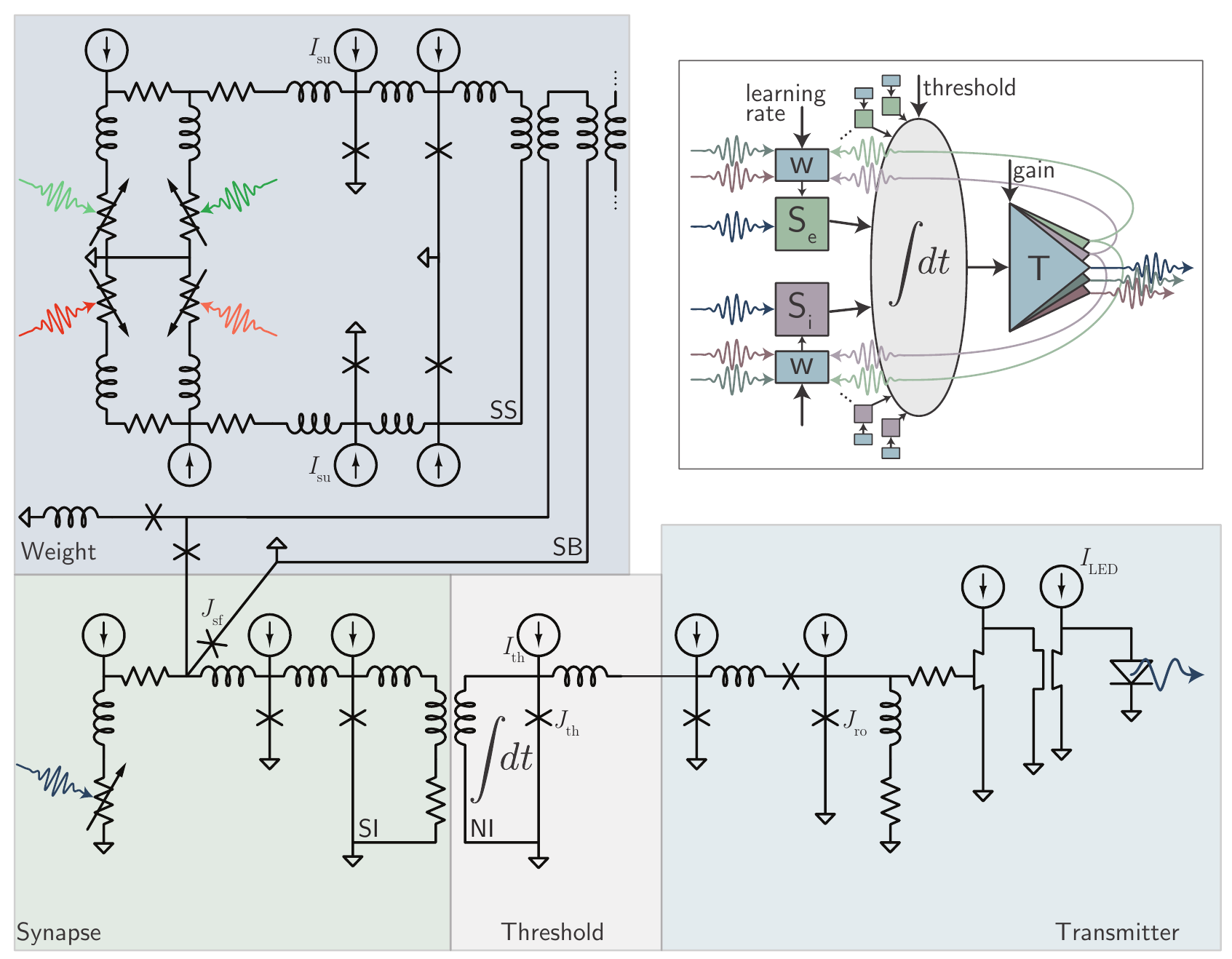}}
	\caption{\label{fig:transmitters_fullCircuit}Circuit diagram of superconducting optoelectronic neuron as described in this series of papers. Large-scale neural systems will employ a wide variety of neurons with myriad structural and dynamical properties. The full potential of neural circuits based on superconducting optoelectronic hardware will only be realized with a long-term experimental and theoretical effort of a broad research community.}
\end{figure*}

With the design of the transmitter portion of the circuit, the neuron model presented in this series of papers is complete. The full circuit is shown in Fig.\,\ref{fig:transmitters_fullCircuit}. The synaptic receiver portion is discussed in Ref.\,\onlinecite{sh2018b}, and the circuit that updates the weights based on spike timing is described in Ref.\,\onlinecite{sh2018c}. The transmitter portion is the subject of the present paper.
Consideration of the circuit within an integrate and fire model is presented in Appendix \ref{apx:IAndF}. 

The synaptic receiver and synaptic weight update circuit in the form presented in Refs.\,\onlinecite{sh2018b} and \onlinecite{sh2018c} utilize separate photons for synaptic firing and synaptic update. In this work, circuit operation culminates by producing light from a single LED, yet it may be desirable to implement transmitters that fire different LEDs for the different operations. This is one way that the power used for syanptic firing events can be decoupled from the power used for synaptic update events. If different colors are used for each of these operations, the same waveguide routing network can be employed for the three signals (synaptic firing, synaptic strengthening, and synaptic weakening), and the demultiplexers located at each downstream neuron can separate the signals and route them locally to the three synaptic ports. Similarly, the spike-timing-dependent plasticity circuit of Ref.\,\onlinecite{sh2018c} requires not only photons from the pre-synaptic neurons, but also photons from the local, post-synaptic neuron. Two additional light sources may be useful at each neuron to be utilized locally for synaptic update. The schematic diagram in Fig.\,\ref{fig:transmitters_schematic} shows these five light sources. While producing five light sources instead of just one adds hardware overhead, the light sources are extremely compact compared to the routing waveguides and inductors associated with the synapses. The synaptic update light sources need not fire nearly as frequently as the synaptic firing light sources. By utilizing independent light sources, it may be possible to reduce network area as well as power consumption. Such operations may be only the beginning of utilizing color in these networks.

In the schematic of Fig.\,\ref{fig:transmitters_fullCircuit}, we emphasize three main current biases that affect the operation of each neuron. A current bias into the synaptic update circuit of Ref.\,\onlinecite{sh2018c} affects the learning rate. A current bias into the neuronal threshold loop \cite{sh2018b} affects the neuron threshold. And a current bias into the light emitter affects the number of photons produced in a neuronal firing event. With each of these currents fixed, the neurons and the network will have rich spatio-temporal dynamics. Yet a network with the ability to dynamically vary these currents will be capable of achieving further complexity over longer time periods. For example, with the learning rate current fixed, spike-timing-dependent plasticity occurs at a fixed rate. By changing this current, the network can make certain regions more adaptive at certain times, and it can make those regions maintain synaptic weights at other times. Similarly, with a fixed gain current into the transmitter, the neuron will address each of its downstream connections with a given probability. By changing the gain current, the number of photons produced in a neuronal firing event can be adjusted, and therefore the probability of reaching downstream connections can be tuned. Changing these bias currents is analogous to changing various neuromodulators in biological systems. The values of each of these neuromodulatory control currents can be modified with photonic or electronic signals. The control currents can also be set externally for experiments in critical dynamics or applications in neural computing, or adjusted based on internal network activity. Adapting these neuromodulatory currents is likely to be useful to achieve maximal information integration through self-organized criticality \cite{be2007,kism2009,shya2009,ch2010,rusp2011}. 

The description of superconducting optoelectronic loop neurons presented in Ref.\,\onlinecite{sh2018b}, Ref.\,\onlinecite{sh2018c}, and in the present work summarizes the operation of these devices in a manifestation that we anticipate being conducive to massive scaling and complex neural operation. The next task is to consider networks comprised of these neurons. Consideration of these networks is presented in Ref.\,\onlinecite{sh2018e}.

\vspace{0.5em}
This is a contribution of NIST, an agency of the US government, not subject to copyright.
	
\newpage
\appendix
	
\section{\label{apx:LEDCircuitModel}LED circuit model}
The equations of motion for this circuit are given by
\begin{equation}
\label{eq:ledDrivenByHTron_equationOfMotion01}
\frac{dI_1}{dt} = \frac{1}{L}\left[V_2+r_1+I_{\mathrm{LED}}-(r_1+r_{\mathrm{hc}})I_1\right];
\end{equation}
\begin{equation}
\label{eq:ledDrivenByHTron_equationOfMotion02}
\frac{dV_2}{dt} = \frac{1}{C}[I_{\mathrm{LED}}-I_1-I_{pn}(V_2)].
\end{equation} 
In these equations, $I_{pn}$ is an analytical function which describes the LED's DC $I-V$ characteristic. It is given by \cite{stba2006}
\begin{equation}
\label{eq:ledDrivenByHTron_ledModel}
\begin{split}
I_{pn} = & eA\left[ (D_p/L_p)p_n + (D_n/L_n)n_p \right] \\
& \times\left[ e^{eV/k_BT} - 1 \right].
\end{split}
\end{equation}
The parameters that enter this equation are as follows. The number of acceptors, $N_a = 5\times 10^{19}$\,cm$^{-3}$. The number of donors, $N_d = 5\times 10^{19}$\,cm$^{-3}$. The intrinsic carrier density, $n_i = 1.5\times 10^{10}$\,cm$^{-3}$. The number of holes on the $n$ side of junction, $p_n = n_i^2/N_d$. The number of electrons on $p$ side of junction, $n_p = n_i^2/N_a$. The electron minority carrier lifetime, $\tau_{np} = 40$\,ns. The hole minority carrier lifetime, $\tau_{pn} = 40$\,ns. The mobility of holes on $p$ side, $\mu_{pp} = 10^{-2}$\,m$^2$/V$\cdot$s. The mobility of holes on $n$ side, $\mu_{pn}= 10^{-2}$ m$^2$/V$\cdot$s. The mobility of electrons on $n$ side, $\mu_{nn} =  2.5\times10^{-2}$\,m$^2$/V$\cdot$s. The mobility of electrons on $p$ side, $\mu_{np} = 2.5\times 10^{-2}$\,m$^2$/V$\cdot$s. The hole diffusion coefficient, $D_p = (k_BT/e)\mu_{pn}$. The electron diffusion coefficient, $D_n = (k_BT/e)\mu_{np}$. The hole diffusion length, $L_p = \sqrt{D_p\tau_{pn}}$. The electron diffusion length, $L_n = \sqrt{D_n\tau_{np}}$. $T = 300$\,K is the temperature of operation. While we plan to operate these circuits at 4.2\,K, our measurements indicate the model captures the operation well at low temperature, and the model is more stable with this room-temperature value. To approximate the capacitance we have assumed the length of the junction is $5$\,\textmu m, the height of the junction is 200\,nm, the intrinsic region is 100 nm, and $\epsilon = 12$ for the semiconductor. This gives $C = 1$\,fF, which is the smallest value we have considered. Fringe fields and wiring parasitics will likely increase this value. We have considered capacitance as large as 100\,fF to ensure our estimates are conservative.
	
\section{\label{apx:hTronCircuitModel}hTron circuit model}
The equations of motion for the circuit in Fig. \ref{fig:transmitters_hTron_circuit}(b) are given by
\begin{equation}
\label{eq:hTron_eq01}
\frac{dT_1}{dt} = \frac{Q}{C_1} + \frac{T_2-T_1}{R_1C_1};
\end{equation}
\begin{equation}
\label{eq:hTron_eq02}
\frac{dT_2}{dt} = \frac{T_1-T_2}{R_1C_2} + \frac{T_3-T_2}{R_2C_2};
\end{equation}
\begin{equation}
\label{eq:hTron_eq03}
\frac{dT_3}{dt} = \frac{T_2-T_3}{R_2C_3} + \frac{T_4-T_3}{R_3C_3};
\end{equation}
\begin{equation}
\label{eq:hTron_eq04}
\frac{dT_4}{dt} = \frac{T_3-T_4}{R_3C_4} + \frac{T_g-T_4}{R_4C_4}.
\end{equation}
In the calculations presented in Fig.\,\ref{fig:transmitters_hTron_data_1} and \ref{fig:transmitters_hTron_data_2}, $Q = I^2r$ where $I = 1.2$\,mA is the current through the gate of the hTron, and $r = 10$\,$\Omega$ is the resistance of the hTron gate. Thus, $Q = 14.4$\,\textmu W.

To produce the plots shown in Fig.\,\ref{fig:transmitters_hTron_data_1} and \ref{fig:transmitters_hTron_data_2}, we model the heater layer as a 10\,nm-thick film of Al. The upper spacer is modeled as a 10 nm-thick film of a-Si, and the lower spacer is modeled as a 50 nm-thick film of SiO$_2$. For the material parameters, the following values are used. The density of Al was taken to be 2700\,kg/m$^3$, and the thermal conductivity was taken to be 30\,W/m$\cdot$K. The density of a-Si was taken to be 2285\,kg/m$^3$, and the thermal conductivity was taken to be 0.01\,W/m$\cdot$K \cite{zipi2006}. The density of SiO$_2$ was taken to be 2650\,kg/m$^3$ and the thermal conductivity was taken to be 0.01\,W/m$\cdot$K. The temperature-dependent specific heat of Al and SiO$_2$ were taken from Ref.\,\onlinecite{du2015}. 

Regarding the superconductor, we anticipate using MoSi for this purpose due to its resistivity in the normal state, its critical current density, and its critical temperature. In a film of 8 nm thickness, the sheet resistance in the normal state is 300 $\Omega/\square$. A wire of 100\,nm width can sustain 16\,\textmu A critical current. A length of 2000 squares achieves 800\,k$\Omega$. A meander of 2000 squares of a wire of 100\,nm width and 50\,nm gap fits in a square of $5.4$\,\textmu m $\times$ 5.4\,\textmu m. At this thickness, $T_c = 6.2$\,K. A device held at 4.2\,K must only be raised by 2\,K to switch to the normal state. The temperature-dependent specific heat of MoSi was taken from Ref.\,\onlinecite{lasa1988}, and the thermal conductivity was taken to be 1\,W/m$\cdot$K.

Of the material parameters, the values of thermal conductivity are the most uncertain. From Ref.\,\onlinecite{zepo1971} we know the thermal conductivity of quartz and silica differ from one another by orders of magnitude at low temperature, and we expect the value of thermal conductivity to depend significantly on the specific material and film deposition method used in fabrication. For functional hTrons fabricated with SiO$_2$ spacer layers and operating between 1 and 5 K, it has been observed \cite{mc2018} that 18\,nW/\textmu m$^2$ is required to switch the hTron. When using the value of 0.01 W/m$\cdot$K for the thermal conductivity of the spacer layers, we arrive at a power density of 494\,nW/\textmu m$^2$, a factor of 27 higher than the measured value. We therefore suspect hTrons can be more efficient than the present model would predict, but further empirical investigation is required to determine the limits of hTron efficiency. The model of Eqs. \ref{eq:hTron_eq01}-\ref{eq:hTron_eq04} neglects phonon reflection at material interfaces. This effect may be responsible for the discrepancy between the model and measurements. Utilizing phonon reflection or phononic crystals in design may further increase the efficiency of the device.
	
\section{\label{apx:hTronInductor}Inductor for the nTron driving the hTron}
Time constants as large as 100\,ns may be necessary for the nTron driving the hTron to produce sufficient photons to serve thousands of synaptic connections. Such a large time constant requires a large inductor, necessitating many squares of meander. The large inductor must also carry a large current (1.2\,mA) to drive the hTron gate, and therefore the wire must be relatively wide. The area of the meander is a concern. With $r_{\mathrm{nT}} = 10$\,$\Omega$, these nTron drive durations require $L_{\mathrm{nT}} = 90$\,nH and $L_{\mathrm{nT}} = 1$\,\textmu H, respectively. The meander achieving this inductance must carry current of 1.2\,mA. If an 8\,nm-thick film of MoSi is utilized for the wire, it will need to be 10\,um wide to carry the current, and it will have 180\,pH/$\square$. In the case of the high-degree neuron producing 10,000 photons per neuronal pulse, 1\,\textmu H inductor meander will occupy 0.6\,mm$^2$. As is shown in Ref.\,\onlinecite{sh2018e}, the area occupied by a neuron of this degree is roughly the same when calculated based on the area of optoelectronic synaptic connections. Thus, while the inductive meander required to achieve the drive duration and to keep the nTron from latching is a large component, because there is only one per neuron (as opposed to one per synapse), it can be straightforwardly accommodated on a dedicated inductive wiring layer. 
	
\section{\label{apx:roParams}Parameters used in threshold circuit}
The $I_c$ of the JJs in the JTL is 250\,\textmu m, and the inductors are 2.1\,pH. The $I_c$ of the JJ just before the relaxation oscillator junction is 280\,\textmu m, and the $I_c$ of the relaxation oscillator junction is the same. The relaxation oscillator is biased with 140\,\textmu A. The value of $L_1 = 200$\,pH . The value of $r_1 = 2.76$\,$\Omega$. The value of $r_2 = 3$\,$\Omega$.

\section{\label{apx:IAndF}Integrate and fire model of loop neuron}
A standard integrate and fire model \cite{daab2001,geki2002} assumes a neuron integrates the signals from all of its synaptic connections, the integrated signal is subject to a leak, the neuron produces a spike when it reaches a threshold, the integrated signal is set to zero upon reaching threshold, and the neuron enters a refractory period after producing a spike during which it cannot produce another spike. Loop neurons operating in this manner can be constructed, yet they may not have the best performance. For loop neurons with synaptic integration loops \cite{sh2018b}, it may be more advantageous for each synaptic integration loop to have its own leak rate. This gives the neuron a much broader range of time constants, contributing to richer dynamics \cite{abre2004,bu2006,be2007}. In this case, a better model treats each synapse with its own integrate and fire equation:
\begin{equation}
\label{eq:IAndF_1}
\frac{dI_i^{(\mathrm{si})}}{dt} = w_i j_i - \frac{I_i^{(\mathrm{si})}}{\tau_i^{(\mathrm{si})}}.
\end{equation}
In Eq. \ref{eq:IAndF_1}, $I_i^{(\mathrm{si})}$ is the current in the synaptic integration loop of the neuron's $i_{\mathrm{th}}$ synapse, $w_i$ is the synaptic weight (expressed as an integer number of fluxons generated furing a synaptic firing event), $\tau_i^{(\mathrm{si})}$ is the $L_{\mathrm{si}}/r_{\mathrm{si}}$ time constant of the synaptic integration loop, and the driving signal into the synapse is given by
\begin{equation}
\label{eq:IAndF_2}
j_i^{(\mathrm{sp})} = j_0\sum_q \delta(t-t_q),
\end{equation}
where $j_0 = \Phi_0/L_i^{(\mathrm{si})}$ is the current added to the synaptic integration loop by a flux quantum, and $t_q$ is the arrival time of the $q_{\mathrm{th}}$ input spike to the synapse. The synapses of loop neurons comprise an SPD in parallel with a JJ. The $L/r$ recovery time of the SPD must be sufficiently long to avoid latching. The required time constant is material dependent and usually in the range of 10 - 50\,ns. During this time, the SPD is biased too low to detect a photon, and the synaptic receiver experiences a refractory period. We therefore take the maximum rate at which synapses can receive signals to be 20\,MHz.

In this model, each synapse is capable of integration and leak, but threshold still occurs for the neuron as a whole. The current in the neuronal integration loop \cite{sh2018b} is a linear combination of the currents in the synaptic integration loops:
\begin{equation}
\label{eq:IAndF_3}
I^{(\mathrm{ni})}(t) = \sum_{i = 1}^{n_{\mathrm{sy}}} c_i I_i^{(\mathrm{si})}(t).
\end{equation}
Here, the coefficients $c_i$ represent the mutual inductors couping the SI loops to the NI loop, and they represent a constant, non-plastic scaling of the synaptic weight.

Firing operation is envisioned as follows. When $I^{(\mathrm{ni})}(t)$ reaches threshold (determined by the thresholding Josephson junction and its dynamic bias), the thresholding junction produces a fluxon that triggers the amplifier chain sequence described in this paper. The neuronal thresholding junction is then induced into a refractory period by the self-feedback nTron circuit of Fig.\,\ref{fig:transmitters_threshold_circuit}(b) (Sec. \ref{sec:refraction}) that diverts the current $I_{\mathrm{th}}$, with exponential recovery. The integrated signals in the synaptic integration loops continue to decay with $L_i^{(\mathrm{si})}/r_i^{(\mathrm{si})}$, but they are not explicitly set to zero. In this mode of operation, the synapses continue to be sensitive to stimulus during the neuron's refractory period. Firing of the neuron does not reset the integrated signal to zero as in the standard integrate and fire formalism. Reset of the integrated currents can be straightforwardly achieved with additional hardware and power, but for many networks it is likely not necessary.

This synaptic leak model can be extended to various levels of dendritic processing with different nonlinearities and time constants present at different levels of dendritic loop hierarchy. Synaptic and dendritic leaks, as opposed to a single neuronal leak, have the advantage of introducing many more decay rates to the repertoire of the neuron. Such time constants may be useful for giving rise to self-organized criticality, as the power law dynamics of systems at the critical point can be constructed as a superposition of exponential functions \cite{fudr2005,be2007}. One disadvantage of synaptic/dendritic leak is the computational overhead in simulation. Each synapse must be represented by an ordinary differential equation in the coupled system. Such a description is more complete \cite{abre2004}, but more expensive to simulate. In the limit that all synaptic time constants are equivalent, one can model the loop neuron with the standard integrate and fire formalism that does not require a separate differential equation for each synapse. 

Eqs.\,\ref{eq:IAndF_1} and \ref{eq:IAndF_3} do not model coincidence detection, sequence detection, or inhibitory dendritic processing, but these effects can be included in more realistic models \cite{geki2002}.

\bibliography{bibliography_modelingSOENs}

\end{document}